\documentclass[letterpaper, 10 pt, conference]{ieeeconf}

\IEEEoverridecommandlockouts                              

\usepackage{graphicx} 
\usepackage{times} 
\usepackage{amsmath} 
\usepackage{amssymb}  
\usepackage{bm}
\graphicspath{{./Figures/}}
\usepackage[caption=false,font=footnotesize]{subfig}
\usepackage{braket}

\usepackage{booktabs,ragged2e}
\usepackage[flushleft]{threeparttable}

\usepackage{xcolor}

\definecolor{orange}{rgb}{1.0, 0.22, 0.0}

\title{\LARGE \bf
Computing a Task-Dependent Grasp Metric Using Second Order Cone Programs}

\author{Amin Fakhari, Aditya Patankar, Jiayin Xie, and Nilanjan Chakraborty
\thanks{The authors are with the Department of Mechanical Engineering, Stony Brook University, Stony Brook, NY 11794, USA, {\tt\small \{amin.fakhari, aditya.patankar, jiayin.xie, nilanjan.chakraborty\}@stonybrook.edu}. This work was supported in part by NSF award CMMI 1853454, and a Stony Brook OVPR Seed Grant.}
}

\begin{document}

\maketitle
\thispagestyle{empty}
\pagestyle{empty}

\begin{abstract}
Evaluating a grasp generated by a set of hand-object contact locations is a key component of many grasp planning algorithms. In this paper, we present a novel second order cone program (SOCP) based optimization formulation for evaluating a grasps' ability to apply wrenches to generate a linear motion along a given direction and/or an angular motion about the given direction. Our quality measure can be computed efficiently, since the SOCP is a convex optimization problem, which can be solved optimally with interior point methods. A key feature of our approach is that we can consider the effect of contact wrenches from any contact of the object with the environment. This is different from the extant literature where only the effect of finger-object contacts is considered. Exploiting the environmental contact is useful in many manipulation scenarios either to enhance the dexterity of simple hands or improve the payload capability of the manipulator. In contrast to most existing approaches, our approach also takes into account the practical constraint that the maximum contact force that can be applied at a finger-object contact can be different for each contact. We can also include the effect of external forces like gravity, as well as the joint torque constraints of the fingers/manipulators. Furthermore, for a given motion path as a constant screw motion or a sequence of constant screw motions, we can discretize the path and compute a global grasp metric to accomplish the whole task with a chosen set of finger-object contact locations.

\end{abstract}

\section{Introduction}
\begin{figure}[!t]
    \centering
    \subfloat[]{\includegraphics[scale=0.35]{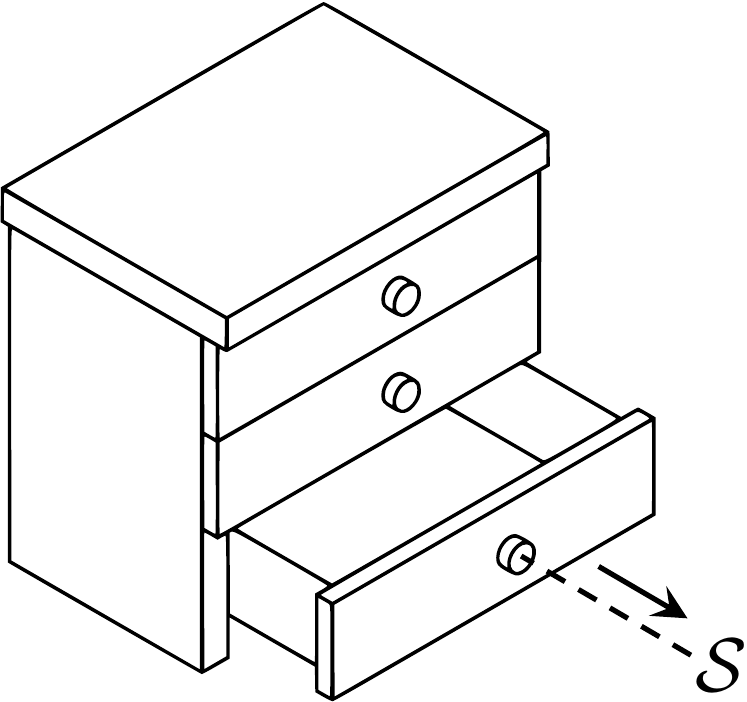}} \hfill
    \subfloat[]{\includegraphics[scale=0.35]{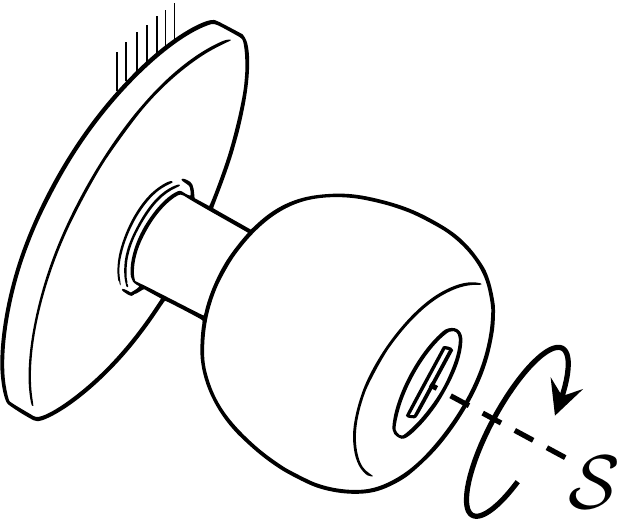}} \hfill
    \subfloat[]{\includegraphics[scale=0.35]{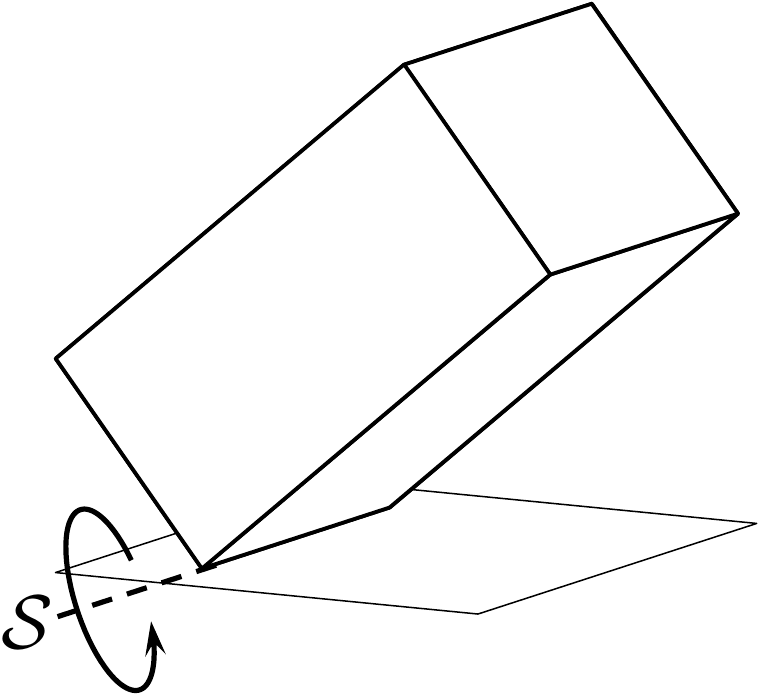}}\\
    \subfloat[]{\includegraphics[scale=0.35]{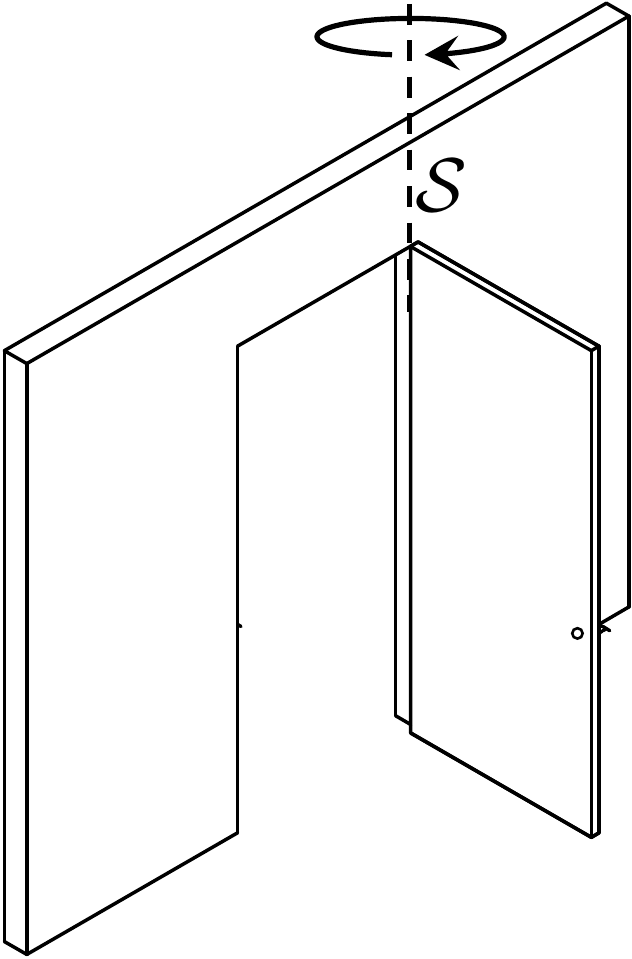}} \hfill
    \subfloat[]{\includegraphics[scale=0.35]{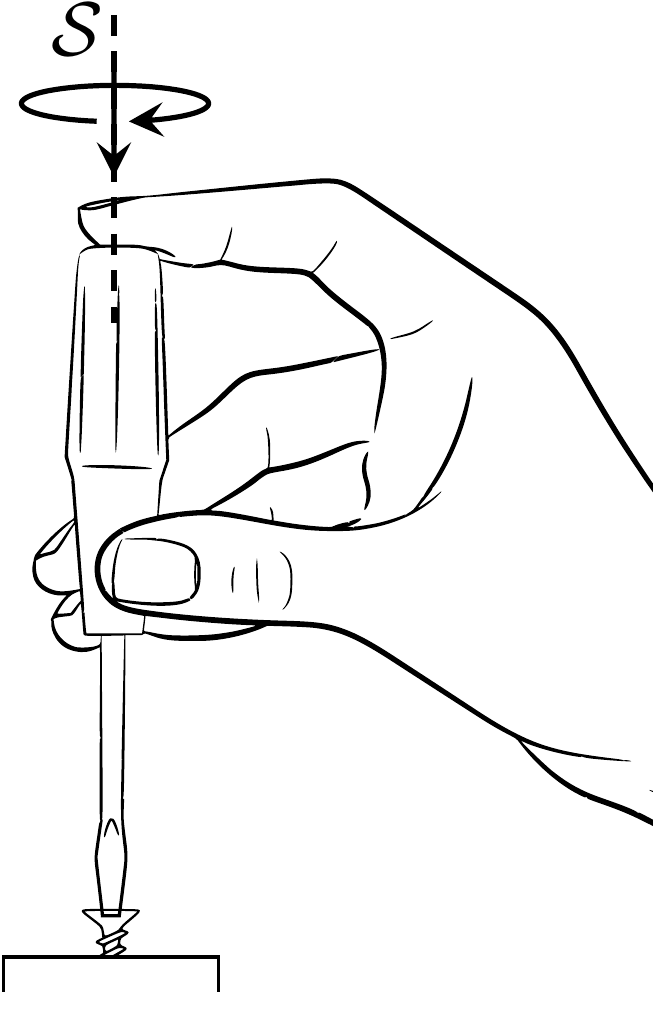}} \hfill
    \subfloat[]{\includegraphics[scale=0.35]{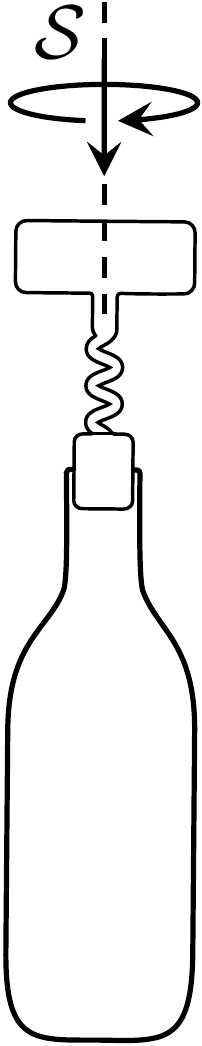}}
    \caption{Examples of performing a task along a screw axis $\mathcal{S}$, (a) pulling open a drawer, (b) turning a door knob, (c) pivoting an object by exploiting the environment, (d) opening/closing a door, (e) fastening/loosening a screw by a screwdriver, (f) twisting a corkscrew.}
\label{Fig:Examples}
\end{figure}

Grasp quality measures or grasp metrics are a key component of many grasp planning algorithms~\cite{miller2003automatic, miller2004graspit, lin2016task, Mahler-RSS-17}. There are usually a multitude of grasps that can be used to hold an object and grasp metrics allow one to quantitatively compare the different grasps. Ideally, a grasp metric or quality measure should reflect the purpose or objective of holding the object. A common purpose of holding an object is to to keep it at a certain pose or pick and place the object. In such situations the maximum amount of disturbance force on the object that the grasp can withstand without losing the object is a natural measure of grasp quality~\cite{FerrariCanny1992}. Another common purpose of holding an object is to generate an instantaneous twist (linear and/or angular velocity) to the object, after grasping it, without losing the grasp.  Figure~\ref{Fig:Examples} shows different manipulation scenarios where the grasp has to generate enough wrench to generate an instantaneous twist. The goal of this paper is {\em to develop a grasp metric that can quantify the ability of a grasp to apply enough wrench to generate an instantaneous twist}. 

In the extant literature grasp metrics that take the finger-object force into account are divided into two main categories, namely, task-independent metrics and task-dependent metrics. Task-independent metrics are useful for the pick and place scenario described above and they seek to measure the maximum amount of disturbance force that the grasp can withstand~\cite{FerrariCanny1992, kirkpatrick1992quantitative}. Task-dependent metrics are useful when the task is characterized by a force/moment that has to be applied to accomplish the task. For example, in Figure~\ref{Fig:Examples} a force and/or a moment has to be applied along or about the axes, $\mathcal{S}$ to perform the task.

The definition of a task-dependent grasp metric is intricately dependent on the definition of a task and there are various definitions of task provided in the literature based on the notion of {\em task wrench space}. 
The task wrench space is defined as a set of wrenches that should be applied on the object along with a set of disturbance wrenches that should be resisted by the grasp while manipulating an object~\cite{RoaSuarez2015,Borst2004, pollard1994parallel}. This set has been geometrically described as an ellipsoid~\cite{li1988task} as well as a polytope~\cite{zhu2001quantitative}. However, it is difficult to accurately obtain a geometric description of task wrench space~\cite{li1988task,krug2017grasp}. In our formulation we define a task as a unit screw, about which the fingers/manipulators need to apply a wrench to generate a desired motion. We will call this unit screw, the {\em task screw}. For example, in Figure~\ref{Fig:Examples}(a), to open the drawer, a force needs to be applied along the direction shown by $\mathcal{S}$. In Figure~\ref{Fig:Examples}(b), (c), and (d) to twist the door knob, to manipulate the object by pivoting, and to open the door respectively, we need to apply a moment about the axis $\mathcal{S}$. In Figure~\ref{Fig:Examples}(e) and (f) a force along $\mathcal{S}$ and simultaneously a moment about $\mathcal{S}$ has to be applied. Our {\em grasp metric is then defined as the maximum magnitude of the wrench that can be applied along the task screw}. This notion of task and grasp metric is similar to that in~\cite{haschke2005task}.


Irrespective of whether the grasp metric is task-dependent or task-independent, a key assumption in the computation of grasp metrics is that either (a) the magnitude of the normal force that can be applied at each robot-object contact is bounded by the same value (usually chosen to be $1$) or (b) concatenating all the normal forces in a vector, some norm (either $1$, $2$ or $\infty$) of the vector is bounded. However as demonstrated experimentally in~\cite{krug2017grasp}, this assumption does not reflect reality. In practice, the maximum amount of force that can be applied at a contact depends on the configuration of the manipulators (or fingers) holding the object as well as the torque limits of the joint motors. This is because, at each contact, the contact force is projected to the joint space through the manipulator (finger) Jacobian, which changes with the configuration. In this paper, we do not assume that the contact wrenches are of unit magnitude and allow the maximum allowable contact force to be different for different contacts. 

{\bf Contributions}: In this paper, we present a novel second order cone program (SOCP) to compute the task-dependent grasp metric defined above. Our formulation reflects practical grasping scenarios by considering independent upper bounds of the normal contact force at each finger-object contact. We  incorporate frictional contact of the object with the environment (if there are any). Thus, our grasp metric can be used to evaluate a grasp meant to manipulate an object by using the environment (which is a useful way for manipulating objects both to increase the dexterity~\cite{dafle2014extrinsic} and payload capability of the robot~\cite{Patankar2020}). To the best of our knowledge, this is the first grasp metric with such capability. We can also consider lower bound constraints on the amount of force transmitted to the support. The lower bound constraints is useful in modeling scenarios like Figure~\ref{Fig:Examples}, where we want to exceed a minimum force along the normal direction between the screwdriver head and the nail. Our formulation also includes the effects of external wrenches (produced by gravity or inertial forces) as well as known external disturbance wrenches. Furthermore, our formulation is flexible enough to include the joint torque constraints explicitly. Since a SOCP is a convex optimization problem, our grasp metric can be efficiently computed using interior point algorithms~\cite{boyd2007fast,boyd2004convex}.

This paper is organized as follows: In Section~\ref{sec:rel_work} we discuss the related work. In Section~\ref{sec:prob}, we present our SOCP formulation for computing the grasp metric. In Section~\ref{sec:res}, we present examples where we show results obtained from our grasp metric and interpret the results. In Section~\ref{sec:conc} we present our conclusions and outline future avenues of work. In the Appendix, we collect some of the terminology that we have used throughout the paper.

\section{Related Work}
\label{sec:rel_work}

Grasp quality metrics have been long recognized as a key component of a large class of grasp synthesis algorithms and therefore has been widely studied. A comprehensive review of grasp quality metrics is given in the excellent survey paper~\cite{RoaSuarez2015}. Therefore, in this paper, we will restrict our attention mostly to the grasp quality metrics that are task-dependent and consider the force applied by the fingers in the evaluation of the metric.

 The concept of wrench spaces has been used extensively to quantify the grasps. The set of all possible wrenches, including the contact wrenches that can be applied at the contacts, is called the \textit{Grasp Wrench Space} (GWS)~\cite{Borst2004, RoaSuarez2015}.  Ferrari and Canny~\cite{FerrariCanny1992} have proposed two grasp quality metrics based on two different ways of approximately computing the GWS assuming a discrete approximation of the friction cone at the finger-object contact. Geometrically, this is equivalent to finding the radius of the largest inscribed sphere centered at the origin and fully contained within the GWS~\cite{FerrariCanny1992,kirkpatrick1992quantitative}. These metrics are task-independent metrics as they do not take into account any information regarding the task. 
 
The concept of this approximate GWS has also been used to take into account the task information and evaluate grasps using a task-related criterion~\cite{lin2016task}. 
In many preceding works, tasks have been defined in terms of a unique set of wrenches acting on the object that should be resisted in order to achieve a specified objective. This is a  convex set and has been described as the \textit{Task Wrench Space} (TWS)~\cite{Borst2004, pollard1994parallel}. The most commonly used method, to evaluate grasps, is to compute the TWS and then scale it to fit the GWS~\cite{Borst2004, pollard1994parallel, RoaSuarez2015, li1988task}. The proposed grasp quality metric is the scaling factor, that can be used to scale the TWS and obtain the largest set that can be fully contained in the GWS. All the metrics described above are task-dependent metrics since they take into account the task-specific TWS while evaluating the grasps. In general, it is very difficult to accurately compute the GWS or the TWS~\cite{RoaSuarez2015,krug2017grasp,lin2016task}.

In~\cite{haschke2005task}, the authors defined a task as a unit wrench in the task wrench space, which is similar to our definition. Although they did not approximate the friction cone constraint, they expressed it as a positive semi-definiteness constraint on a matrix and the problem of computing the grasp metric has been formulated using Linear Matrix Inequalities (LMI)~\cite{buss1996dexterous}. The LMI formulation embeds the friction cone constraint in a higher dimensional space and the size of the problem increases and the computation cost increases. This has been noted in~\cite{boyd2007fast}, in the context of grasp analysis problems, in which the authors demonstrated that using a SOCP formulation leads to much faster algorithms compared to the LMI formulations that were used extensively~\cite{han2000grasp,helmke2002quadratically,buss1996dexterous}. Therefore, in this paper we model the friction cones as a second order cone.

As discussed before, all of these methods of computing grasp metrics, whether task-dependent or task-independent, do not reflect the practical situation that the normal force at each finger object contact can be bounded by a different maximum value that depends on the hand configuration and joint torque limits. Krug \textit{et. al} studied the effects of these underlying assumptions in~\cite{krug2017grasp} while evaluating various grasps and found that they do not accurately reflect the quality of the grasp. Therefore, we do not assume the magnitude of independent contact wrenches or the sum of the magnitude of the contact wrenches to be upper-bounded by $1$. Our formulation allows for contact wrenches to have different magnitudes at different contact locations. We can also explicitly take into account the joint torque constraints of each finger joint.

Recent works on task-based metrics have characterized a task by a single task-specific wrench acting along a particular direction in the TWS \cite{haschke2005task, krug2016analytic, krug2018evaluating}. In some cases, both the magnitude and the direction of the task-specifc wrench is of importance~\cite{song2020robust}. However, none of the methods discussed thus far take into consideration the object-environment contact which is critical in many manipulation tasks. For measuring grasp qualities for tasks shown in Figure.~\ref{Fig:Examples}(c) or Figure.~\ref{Fig:Examples}(e), the object-environment contacts has to be taken into consideration as they are an inherent part of the task being executed. Our SOCP based formulation not only considers the object-manipulator contact wrenches but also the reaction forces at the object-environment contact.

\section{Task-Dependent Grasp Metric}
\label{sec:prob}
\begin{figure}[!htbp]
\centering
\includegraphics[scale=0.5]{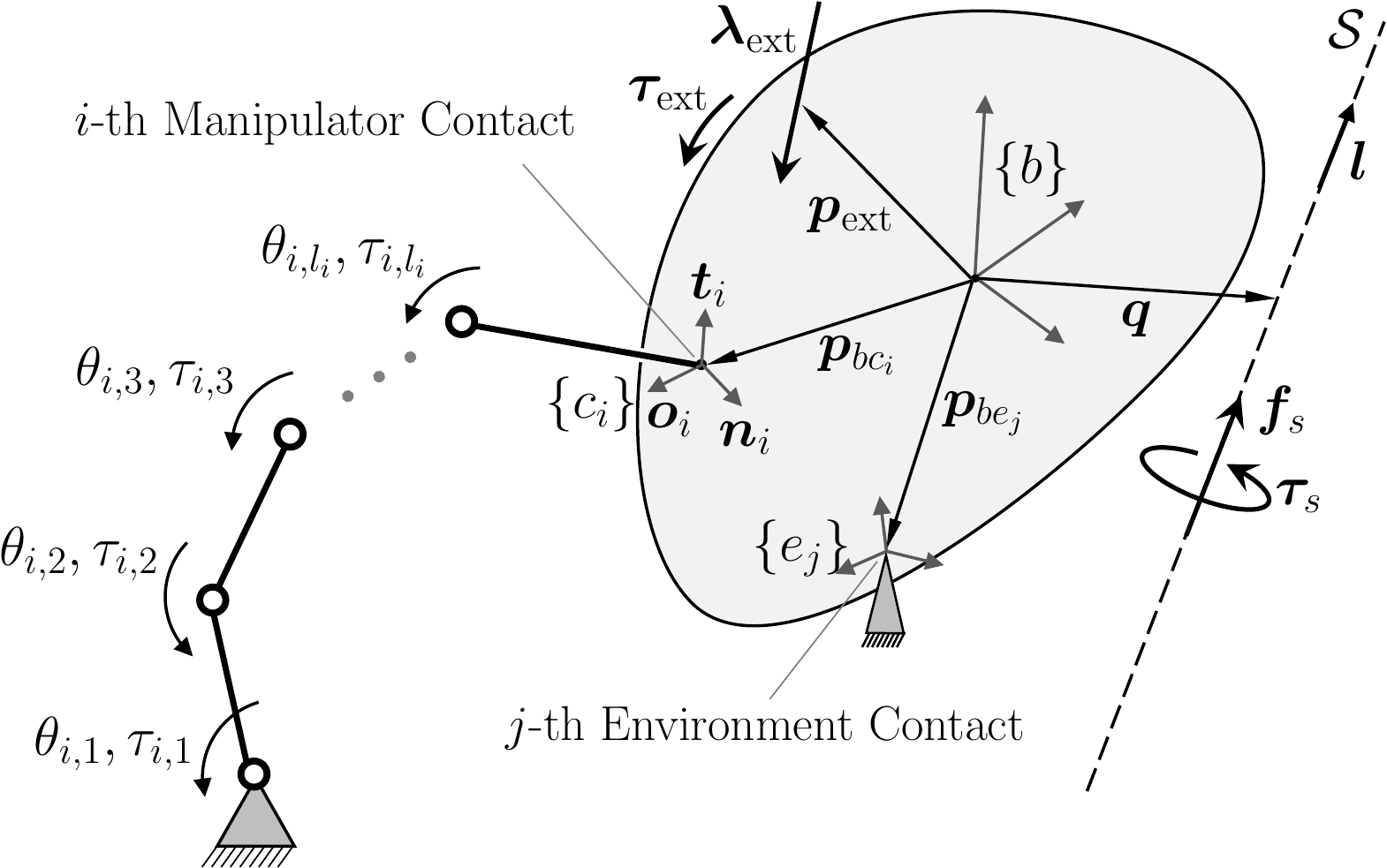}
\caption{A grasped object performing a task along the screw axis $\mathcal{S}$. }
\label{Fig:Object_Manipulator}
\end{figure}

Consider a rigid object which is grasped by $n$ manipulators (or fingers) at $n$ contact positions, $c_i$, ($i=1,\dots,n$) to perform a task as shown in Fig.~\ref{Fig:Object_Manipulator}. Let $e_j$ ($j=1,\dots,m$) be the object environment contact points. Let $\mathcal{S}$ be the screw axis about which we need to apply a wrench. $\mathcal{S} = (\bm{l}, \bm{q})$, where $\bm{l}$ is a unit vector representing a direction and $\bm{q}$ is a point on the screw axis $\mathcal{S}$.
The contact frame \{$c_i$\} is attached to the object at each manipulator contact, such that the axis of the frame ${\bm n}_i$ is normal (inward) to the contact surface and two other axes, ${\bm t}_i$, and ${\bm o}_i$, are tangent to the object surface. 

\subsection{Object-Manipulator Contacts Wrenches}
In this paper, it is assumed that the contacts between the grasped object and the manipulators are soft finger contact with elliptic approximation (SFCE). Thus, the wrench  at each contact can be expressed, in the contact frame $\{c_i\}$, as $\bm{f}_{c_i}=[{f}_{c_{t,i}}, {f}_{c_{o,i}}, {f}_{c_{n,i}}, 0, 0,  {m}_{c_{n,i}}]^\mathrm{T} \in \mathbb{R}^6$ ($i=1,\dots,n$) where ${f}_{c_{t,i}}$ and ${f}_{c_{o,i}}$ are the tangential frictional forces, ${f}_{c_{n,i}}$ is the normal force, and ${m}_{c_{n,i}}$ is the frictional moment. The components of $\bm{f}_{c_i}$ should satisfy an ellipsoidal constraint \begin{equation}
  \frac{1}{{\mu_c}_i} \sqrt {{\left( {\frac{{f}_{c_{t,i}}}{{e}_{c_{t,i}}}} \right)^2} + {\left( {\frac{{f}_{c_{o,i}}}{{e}_{c_{o,i}}}} \right)^2} + {\left( {\frac{{m}_{c_{n,i}}}{{e}_{c_{n,i}}}} \right)^2}} \le {f}_{c_{n,i}},
  \label{equation:SFCE}
\end{equation}
where ${\mu_c}_i$, ${e}_{c_{t,i}}$, ${e}_{c_{o,i}}$, ${e}_{c_{n ,i}} \in \mathbb{R}_{++}$ are the parameters of anisotropic friction at the $i$-th manipulator contact. This constraint is a \textit{Second-Order Cone} (SOC) constraint in the form
\begin{equation}
  {\mathcal{K}_c}_i = \left\{ {\bm x} \;\middle|\; \frac{1}{{\mu_c}_i} \sqrt { {\frac{x_1^2}{{e}_{c_{t,i}}^2} } + { {\frac{x_2^2}{{e}_{c_{o,i}}^2}} } + { {\frac{x_6^2}{{e}_{c_{n,i}}^2}} }} \le x_3, x_{4,5}=0 \right\},
\end{equation}
where ${\bm x} \in \mathbb{R}^6$~\cite{boyd2007fast}. Therefore, the constraint \eqref{equation:SFCE} can be concisely represented by the friction cone ${\mathcal{K}_c}_i$ as
\begin{equation}
{\bm f}_{c_i} \in {\mathcal{K}_c}_i, \quad i=1, \dots, n.
\label{equation:SFCE_compact_1}
\end{equation}
By concatenating all the contact wrenches ${\bm f}_{c_i}$ as ${\bm f}_{c}= [{\bm f}_{c_1}^\mathrm{T},\dots,{\bm f}_{c_n}^\mathrm{T}]^\mathrm{T} \in \mathbb{R}^{6n}$ and all the friction cones sets ${\mathcal{K}_c}_i$ as ${\mathcal{K}_c} = \left\{ {{\bm x}} \in \mathbb{R}^{6n} \;\middle|\; {{\bm x}}_i \in {\mathcal{K}_c}_i,\,\,  i=1, \dots, n \right\}$
the constraint \eqref{equation:SFCE_compact_1} can be rewritten as
\begin{equation}
{\bm f}_{c} \in {\mathcal{K}_c}.
\label{equation:SFCE_compact_2}
\end{equation}

If the grasped object is breakable, we should also have a constraint on the maximum normal force exerted by each of the manipulators on the object as
\begin{equation}
{\bm f}_{n} \leq {\bm f}_{n,\mathrm{max}},
\label{equation:normal_force_constraint}
\end{equation}
where ${\bm f}_{n} = [{f}_{c_{n,1}}, \dots, {f}_{c_{n,n}}]^\mathrm{T} \in \mathbb{R}^n$ is the vector of normal contact forces, and ${\bm f}_{n,\mathrm{max}} \in \mathbb{R}^n$ is a positive vector that represent the upper limit for these forces. 

In order to find the total wrench exerted by the manipulators through the contacts, we should first express all contact wrenches in the same frame. By choosing $\{b\}$ as the fixed body frame, the wrench ${\bm f}_{c_i}$ can be expressed in this frame by means of the adjoint matrix ${\mathbf{G}}_{c_i} \in \mathbb{R}^{6 \times 6}$ as
\begin{equation}
     {\bm f}_{c_i}^b = {\mathbf{G}}_{c_i} {\bm f}_{c_i} = \begin{bmatrix} {\mathbf{R}_{bc_i}} & {\bm 0} \\ ({\bm p}_{bc_i})_{\times} {\mathbf{R}_{bc_i}} &  {\mathbf{R}_{bc_i}}  \end{bmatrix} {\bm f}_{c_i},
\end{equation}
where ${\mathbf{R}_{bc_i}} \in SO(3)$ is the $3\times3$ rotation matrix of frame $\{c_i\}$ with respect to frame $\{b\}$, ${\bm p}_{bc_i} \in \mathbb{R}^3$ is the positions of $\{c_i\}$ with respect to $\{b\}$ and expressed in $\{b\}$, and $(\cdot)_{\times}$ is the $3 \times 3$ skew-symmetric matrix representation of a vector. By defining the concatenated matrix 
$\mathbf{G}_{c} = [{{\bm{\mathrm{G}}}}_{c_1}, \dots, {{\bm{\mathrm{G}}}}_{c_n}] \in \mathbb{R}^{6 \times 6n}$, the total wrench the manipulators can apply to the grasped object is 
\begin{equation}
    {\bm f}_{c}^b = \sum_{i = 1}^{n}{{\bm f}_{c_i}^b} = {{{\bm{\mathrm{G}}}}_{c} {\bm f}_{c}}.
\end{equation}

\subsection{Object-Environment Contacts Wrenches}
When the object is in contact with the environment, we should also consider the wrenches applied to the object through these environment contacts. Let the vector ${\bm f}_{e_j} \in \mathbb{R}^6$ ($j=1,\dots,m$) be the wrench applied to the object at the environment contact $\{e_i\}$. If this wrench is expressed in the frame $\{e_i\}$, the total wrench, expressed in $\{b\}$, that the environment can apply to the object is computed as
\begin{equation}
    {\bm f}_{e}^b = \sum_{j = 1}^{m}{{\mathbf{G}}_{e_j}} {\bm f}_{e_j} = \sum_{j = 1}^{m}{\begin{bmatrix} {\mathbf{R}_{be_j}} & {\bm 0} \\ ({\bm p}_{be_j})_{\times} {\mathbf{R}_{be_j}} &  {\mathbf{R}_{be_j}}  \end{bmatrix} {\bm f}_{e_j}},
    \label{eq:ExternalWrenches}
\end{equation}
where ${\mathbf{G}}_{e_j} \in \mathbb{R}^{6 \times 6}$ is the adjoint matrix, ${\mathbf{R}_{be_j}} \in SO(3)$ is the $3\times3$ rotation matrix of frame $\{e_j\}$ with respect to frame $\{b\}$, ${\bm p}_{be_j} \in \mathbb{R}^3$ is the positions of $\{e_j\}$ with respect to $\{b\}$ and expressed in $\{b\}$. By defining the concatenated matrix $\mathbf{G}_{e} = [{\bm{\mathrm{G}}}_{e_1}, \dots, {\bm{\mathrm{G}}}_{e_m}] \in \mathbb{R}^{6 \times 6m}$, and the concatenated vector ${\bm f}_{e} = [{\bm f}_{e_1}^\mathrm{T},\dots,{\bm f}_{e_m}^\mathrm{T}]^\mathrm{T} \in \mathbb{R}^{6m}$, \eqref{eq:ExternalWrenches} can be rewritten as
\begin{equation}
    {\bm f}_{e}^b = {{{\bm{\mathrm{G}}}}_{e} {\bm f}_{e}}.
\end{equation}


\subsection{External Wrenches}
Let $\bm{\lambda}_{\mathrm{ext}} \in \mathbb{R}^3$ and $\bm{\tau}_{\mathrm{ext}} \in \mathbb{R}^3$ be the total external force and moment applied to the object (including the object weight), expressed in the fixed body frame $\{b\}$, respectively. Thus, the total external wrench ${\bm f}_{\mathrm{ext}} \in \mathbb{R}^6$ applied to the object in frame $\{b\}$ is
\begin{equation}
    {\bm f}_{\mathrm{ext}} = \left[ \begin{array}{c}
	\bm{\lambda}_{\mathrm{ext}}\\
	{\bm p}_{\mathrm{ext}} \times \bm{\lambda}_{\mathrm{ext}} + \bm{\tau}_{\mathrm{ext}}\\
\end{array} \right],
\end{equation}
where ${\bm p}_{\mathrm{ext}}$ is  a vector from the origin of the frame $\{b\}$ to the line of action of the external force $\bm{\lambda}_{\mathrm{ext}}$.

\subsection{Manipulator Joint Torque Constraints}
By defining the vectors of joint variables and joint torques as ${\bm \theta}_i = [\theta_{i,1}, \dots, \theta_{i,l_i}]^\mathrm{T}$ and $\bm{\tau}_i = [\tau_{i,1}, \dots, \tau_{i,l_i}]^\mathrm{T}$, respectively, for the serial, fully-actuated $i$-th $l_i$-DOF manipulator, the relationship between the joint torques $\bm{\tau}_i$ and the contact wrench ${\bm f}_{c_i}$ is
\begin{equation}
  \bm{\tau}_i = - {{\bm{\mathrm{J}}}}_i^\mathrm{T}({\bm \theta}_i) {\bm f}_{c_i} + {\bm{\tau}_g}_i({\bm \theta}_i), \quad i=1, \dots, n,
  \label{equation:joint_torques}
\end{equation}
where ${\bm{\mathrm{J}}}_i({\bm \theta}_i) \in \mathbb{R}^{6 \times l_i}$ is the Jacobian matrix expressed in the contact frames \{$c_i$\} and ${\bm{\tau}_g}_i({\bm \theta}_i) \in \mathbb{R}^ {l_i}$ is the vector of joint torques due to gravity. Equation \eqref{equation:joint_torques} can be concisely rewritten for all $n$ manipulators as
\begin{equation}
  \bm{\tau} = - {\bm{\mathrm{J}}}^\mathrm{T}({\bm \theta}) {\bm f}_{c} + \bm{\tau}_g(\bm{\theta}),
  \label{equation:joint_torques_compact}
\end{equation}
where $\bm{\theta} = [\bm{\theta}_1^\mathrm{T},..., \bm{\theta}_n^\mathrm{T}]^\mathrm{T} \in \mathbb{R}^{l}$, $\bm{\tau} = [\bm{\tau}_1^\mathrm{T},..., \bm{\tau}_n^\mathrm{T}]^\mathrm{T} \in \mathbb{R}^{l}$, ${\bm{\tau}_g} = [{\bm{\tau}_g}_1^\mathrm{T}, \dots, {\bm{\tau}_g}_n^\mathrm{T}]^\mathrm{T} \in \mathbb{R}^{l}$, ${\bm{\mathrm{J}}} = \mathrm{diag}({\bm{\mathrm{J}}}_1, \dots, {\bm{\mathrm{J}}}_n) \in \mathbb{R}^{6n \times l}$, and $l = \sum_{i = 1}^{n}{l_i}$. Let the upper and the lower limits of the manipulators joint torques be $\bm{\tau}_{\rm max} \in \mathbb{R}^{l}$ and $\bm{\tau}_{\rm min} \in \mathbb{R}^{l}$, respectively. Therefore, the joint torque constraints can be represented as
\begin{equation}
    \bm{\tau}_\mathrm{min} \leq \bm{\tau} \leq \bm{\tau}_\mathrm{max}.
    \label{equation:joint_torques_compact_limits}
\end{equation}

\subsection{Grasp Wrench Space}
We define the Grasp Wrench Space (GWS) as a set of all possible wrenches that a grasp, including the manipulators, environment, and external wrenches, can apply to an object by considering all the wrench/contact constraints. Mathematically, this set can be represented as
\begin{equation}
  \mathcal{W} = \left\{\bm{w} \;\middle|\; \bm{w} =  {{\bm{\mathrm{G}}}_{c} \bm{f}_{c}} + {{\bm{\mathrm{G}}}_{e} \bm{f}_{e}} + \bm{f}_\mathrm{ext}, \, \bm{f}_{c} \in {\mathcal{B}_c}, \, \bm{f}_{e} \in {\mathcal{B}_e}\right\},
  \label{eq:GWS}
\end{equation}
where $\bm{w} \in \mathbb{R}^6$, $\mathcal{B}_c$ is a set of all the manipulator contact wrenches which satisfy the contact constraints in \eqref{equation:SFCE_compact_2} and \eqref{equation:normal_force_constraint} as well as manipulator joint torque constraints in \eqref{equation:joint_torques_compact_limits}, and $\mathcal{B}_e$ is a set of all the object-environment contacts wrenches which satisfy the environment contact constraints. These constraints can be friction cone constraints or wrench magnitude upper/lower bound for frictional or fixed contacts. Notice that this six-dimensional (6D) space is convex and includes the origin $\bm{0}$. 

The definition of grasp wrench space provided above is different from the usual definitions of grasp wrench space in the extant literature. The grasp wrench space was defined as the convex hull of the Minkowski sum of the wrenches produced by the forces on the boundary of the discretized friction cones at the contact points, while assuming all contact forces have the same upper limit normalized to $1$ \cite{FerrariCanny1992}. This definition was later expressed as $\mathcal{W} = \left\{\bm{w} \in \mathbb{R}^6 \;\middle|\; \bm{w} =  {{\bm{\mathrm{G}}}_{c} \bm{f}_{c}}, \, \bm{f}_{c} \in {\mathcal{K}_c}, \lVert \bm{f}_{c} \rVert \leq 1 \right\}$, where the norm could be the $1$, $2$, or infinity norm~\cite{Borst2004,haschke2005task,RoaSuarez2015}. However, in general, the force upper limit at all the contacts may not be the same. Moreover, environment contacts and external wrenches may be also present in a grasp. Thus, \eqref{eq:GWS} represents a more general definition of the grasp wrench space. One reason for having the general definition is that the external wrenches may also assist (not only impede) in performing the task.

\subsection{Task-Dependent Grasp Metric as a SOCP}
According to Poinsot’s theorem, every wrench applied to a rigid body is equivalent to a force applied along a fixed screw axis and a torque about the same axis \cite{Murray1994}. Since we define a task as a constant screw motion/wrench along a given axis $\mathcal{S}$, we are interested in finding a wrench $\bm{w}_{\mathrm{task}} \in \mathcal{W}$, among all the possible wrenches that a grasp can generate, which is equivalent to a wrench along the screw axis $\mathcal{S}$.


By defining $\bm{l} \in \mathbb{R}^3$ as a unit vector along the screw axis $\mathcal{S}$ (Fig.~\ref{Fig:Object_Manipulator}), we can equivalently represent the wrench $\bm{w}_{\mathrm{task}}$ by a force $\bm{f}_s \in \mathbb{R}^3$ along the axis $\mathcal{S}$ (i.e., $\bm{f}_s = f_s \bm{l}$) and a torque $\bm{\tau}_s \in \mathbb{R}^3$ about this axis (i.e., $\bm{\tau}_s = \tau_s \bm{l}$) as

\begin{equation}
\bm{w}_{\mathrm{task}} = \begin{bmatrix} \bm{f}_t \\
\bm{\tau}_t \end{bmatrix} = \begin{cases}
	f_s \begin{bmatrix} \boldsymbol{l} \\
\bm{q} \times \bm{l} + h\bm{l}\end{bmatrix}, &		h\in \mathbb{R}, \\[5mm]
	\tau_s \begin{bmatrix} \boldsymbol{0} \\
\bm{l} \end{bmatrix}, &		h=\infty,
\end{cases}
\label{eq:w_task}
\end{equation}
where $h = \tau_s/ f_s \in \mathbb{R}$ is known as the screw pitch and can be computed by $h = \bm{f}_t^\mathrm{T} \bm{\tau}_t / {\lVert \bm{f}_t \rVert}^2$, and $\bm{q} \in \mathbb{R}^3$ is any point on the screw axis $\mathcal{S}$ and can be found by  $\bm{q} = \bm{f}_t \times \bm{\tau}_t / {\lVert \bm{f}_t \rVert}^2 + \lambda \bm{l}$ ($\forall\lambda \in \mathbb{R}$). When $\bm{w}_{\mathrm{task}}$ has only a pure torque $\bm{\tau}_t$ (i.e., $\bm{f}_t = \bm{0}$), $h=\infty$ and the screw axis $\mathcal{S}$ is along $\bm{\tau}_t$, otherwise, it is along $\bm{f}_t$. Note that the term $ \bm{q} \times f_s\bm{l}$ in \eqref{eq:w_task} represents the moment of the force $\bm{f}_s$ about the origin of the reference frame. Therefore, any spatial wrench $\bm{w}_{\mathrm{task}}$ can be represented by three screw coordinates of the magnitude ($f_s$ or $\tau_s$), the pitch ($h$), and the axis ($\bm{l}$ and $\bm{q}$). 

\begin{figure}[!htbp]
\centering
\includegraphics[scale=0.35]{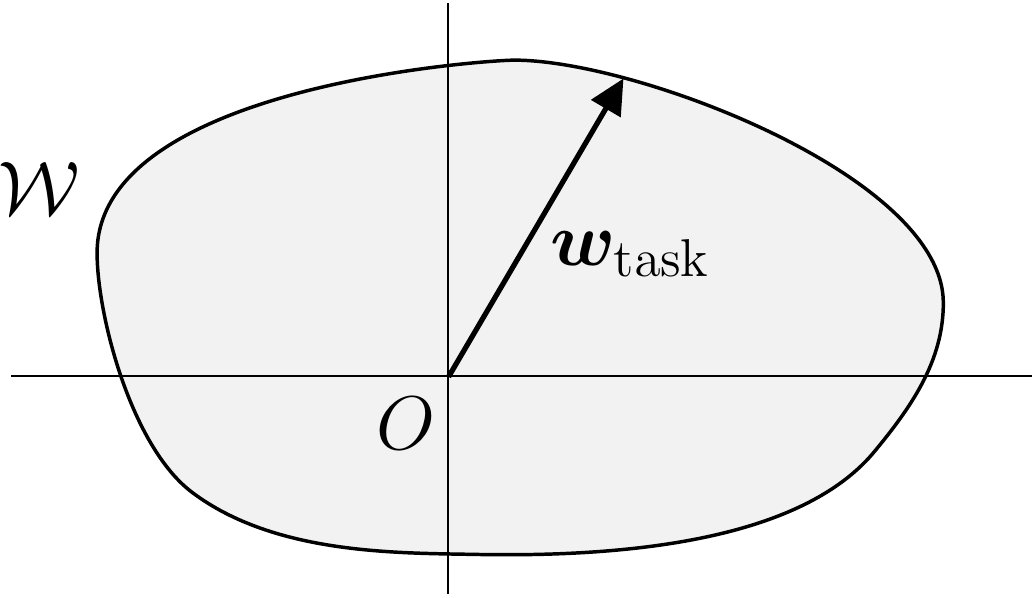}
\caption{A wrench with maximum magnitude along a given screw axis $\mathcal{S}$ in grasp wrench space $\mathcal{W}$.}
\label{Fig:GWS}
\end{figure}

Hence, by representing a task as a constant screw motion/wrench along a given axis $\mathcal{S}$ ($\bm{l}$ and $\bm{q}$) with a given pitch $h$, we define \textit{task-dependent grasp metric} $\eta$ as the maximum magnitude of a wrench that the grasp can generate along a given screw axis $\mathcal{S}$ among all the possible wrenches in the GWS $\mathcal{W}$ (Fig.~\ref{Fig:GWS}).
Mathematically, we can compute the metric $\eta$ by defining a second order cone program (SOCP) as
\begin{equation}
\begin{aligned}
& {\underset {\bm{f}_{c},\, \bm{f}_{e},\, \bm{\tau},\, \eta}{\text{maximize} }} & & \eta \\
& \text{subject to}  & &  {{\bm{\mathrm{G}}}_{c} \bm{f}_{c}} + {{\bm{\mathrm{G}}}_{e} \bm{f}_{e}} + \bm{f}_\mathrm{ext} =  \bm{w}_{\mathrm{task}}, \\
&&& \bm{f}_{c} \in {\mathcal{K}_c},\\
&&& \bm{f}_{n} \leq \bm{f}_{n,\mathrm{max}}, \\
&&& \bm{\tau} + {\bm{\mathrm{J}}}^\mathrm{T} \bm{f}_{c} - \bm{\tau}_g = \bm{0},\\
&&& \bm{\tau}_{\rm min} \leq \bm{\tau} \leq \bm{\tau}_{\rm max},\\
&&& \bm{f}_{e} \in {\mathcal{B}_e},
\end{aligned}
\label{equation:updated_optimization}
\end{equation}
where $\eta = f_s$ when $h \in \mathbb{R}$, $\eta = \tau_s$ when $h=\infty$, and the vectors $\bm{l}$ and $\bm{q}$ are expressed in the frame $\{b\}$. In \eqref{equation:updated_optimization}, $\bm{f}_{c}$, $\bm{f}_{e}$, $\bm{\tau}$, and $\eta$ are the optimization variables, and $\bm{l}$, $\bm{q}$, and $h$ are given for a specific task. Therefore, the metric $\eta$, computed from the optimization problem \eqref{equation:updated_optimization}, determines the quality of the grasp for performing a specific task along a specific axis.

The grasp metric present above is for a particular instant. However, we can use this formulation to compute a grasp metric during the whole task.
The grasp metric $\eta$ will usually not be constant along the whole task/motion. This can be due to many reasons including (i) change in the direction or magnitude of the environment wrench ${{{\bm{\mathrm{G}}}}_{e} {\bm f}_{e}}$ and/or external wrench ${\bm f}_{\mathrm{ext}}$ applied to the object while performing a task along a fixed screw axis (it can be because of the effect of the object weight, springs, variable friction coefficients, etc.), (ii) defining a complex task as a sequence of constant screw motions about different screw axes, (iii) presence of inertial forces in non-quasi-static motions. One of the advantages of our proposed method is that it can deal with all these situations. Indeed, we can find a \textit{local} metric $\eta$ at discrete poses on the motion path using \eqref{equation:updated_optimization} and then, obtain a \textit{global} metric $\eta^*$ over the entire task by finding the minimum value among the local metrics. This metric can be used with the existing grasp planning algorithms to compute the best possible grasp over the whole motion or task.


\section{Implementation and Results}
\label{sec:res}
In this section we compute the proposed grasp metric for three different tasks of turning a door handle, pivoting an object, and sliding an object on a support surface. In these examples, for simplicity, we do not consider the manipulators joint torque constraint \eqref{equation:joint_torques_compact_limits}. Since the problem \eqref{equation:updated_optimization} is a convex optimization problem, the following simulations have been implemented using the CVX toolbox \cite{cvx} in MATLAB with  the default solver (SDPT3) on a 3.00 GHz quad-core processor and 8 GB RAM.  

\subsection{Turning a Door Handle}
\begin{figure}[!htbp]
\centering
\includegraphics[scale=0.5]{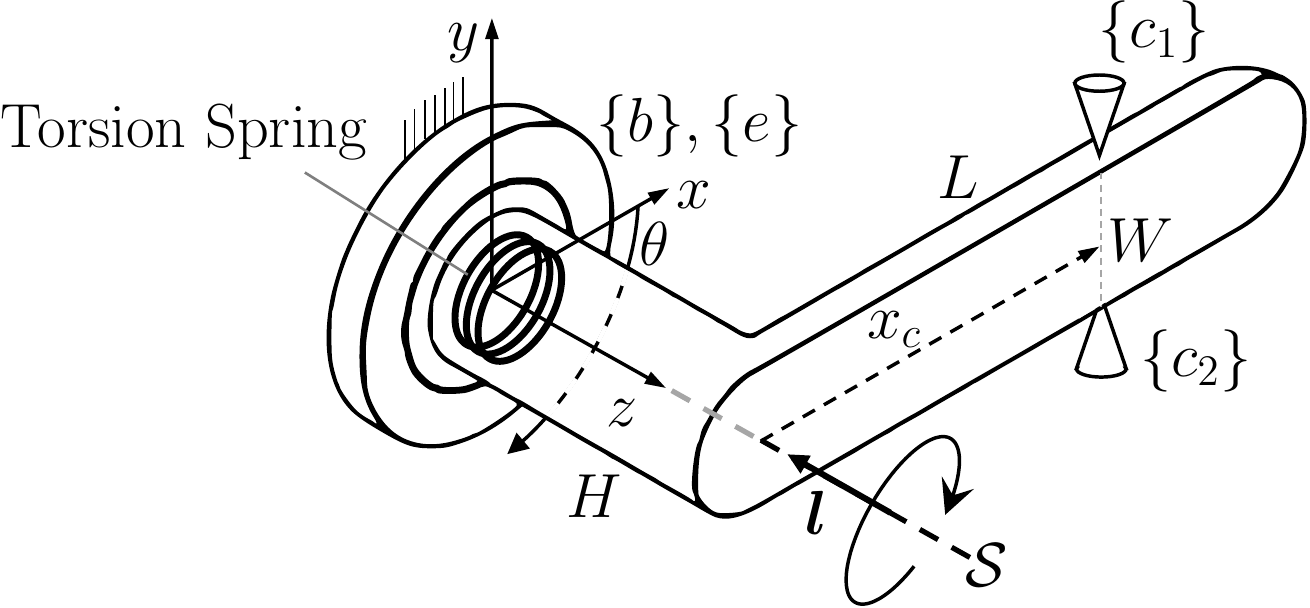}
\caption{Turning a door handle by a two-finger antipodal grasp.}
\label{Fig:DoorHandle}
\end{figure}

In this example, we evaluate the grasp metric $\eta$ while turning a door handle by a pair of antipodal contact points as shown in Fig.~\ref{Fig:DoorHandle}.
The manipulator contacts ($\{c_1\}$, $\{c_2\}$) are considered to be soft finger contact with elliptic approximation (SFCE), the body frame $\{b\}$ is attached to the handle at the fixed support, the handle weight is assumed to be negligible,
and torque of the the torsion spring at the hinge of the door handle is $k_t \theta$ where $k_t \in \mathbb{R}$ is the spring constant and $\theta \in \mathbb{R}$ is angle of the handle from its horizontal position (Fig.~\ref{Fig:DoorHandle}).
Thus, the wrench applied to the handle at the environment contact $\{e\}$ (the fixed support) is $\bm{f}_{e}^b=[{f}_{e_x}, {f}_{e_y}, {f}_{e_z}, {m}_{e_x}, {m}_{e_y}, k_t \theta]^\mathrm{T} \in \mathbb{R}^6$.
The simulation parameters are given in Table~\ref{Table:DoorHandle_parameters}, moreover, it is assumed that the maximum normal force at each manipulator contact is ${f}_{c_{n,1}} = {f}_{c_{n,2}} = 20$ N, and $k_t = 0.6$ Nm/rad.

\begin{table}[!htbp]
    \caption{Simulation parameters for turning the door handle.}
    \centering
    \setlength{\tabcolsep}{1mm}
    \renewcommand{\arraystretch}{1.2}
    \begin{tabular}{c|c}
        \hline
        \textbf{Parameter}&\textbf{Value}  \\
        \hline 
        \hline 
        Dimensions & $L = 0.20$ (m), $H = 0.04$ (m), $W = 0.03$ (m)\\
        \hline
        Constants & ${\mu_c}_{1,2}=0.20$, ${e}_{c_{n,1,2}}=0.03$ (m), ${e}_{c_{t,1,2}}={e}_{c_{o,1,2}}=1$  \\
        \hline
    \end{tabular}
    \label{Table:DoorHandle_parameters}
\end{table}

\begin{figure}[!htbp]
\centering
\includegraphics[scale=0.55]{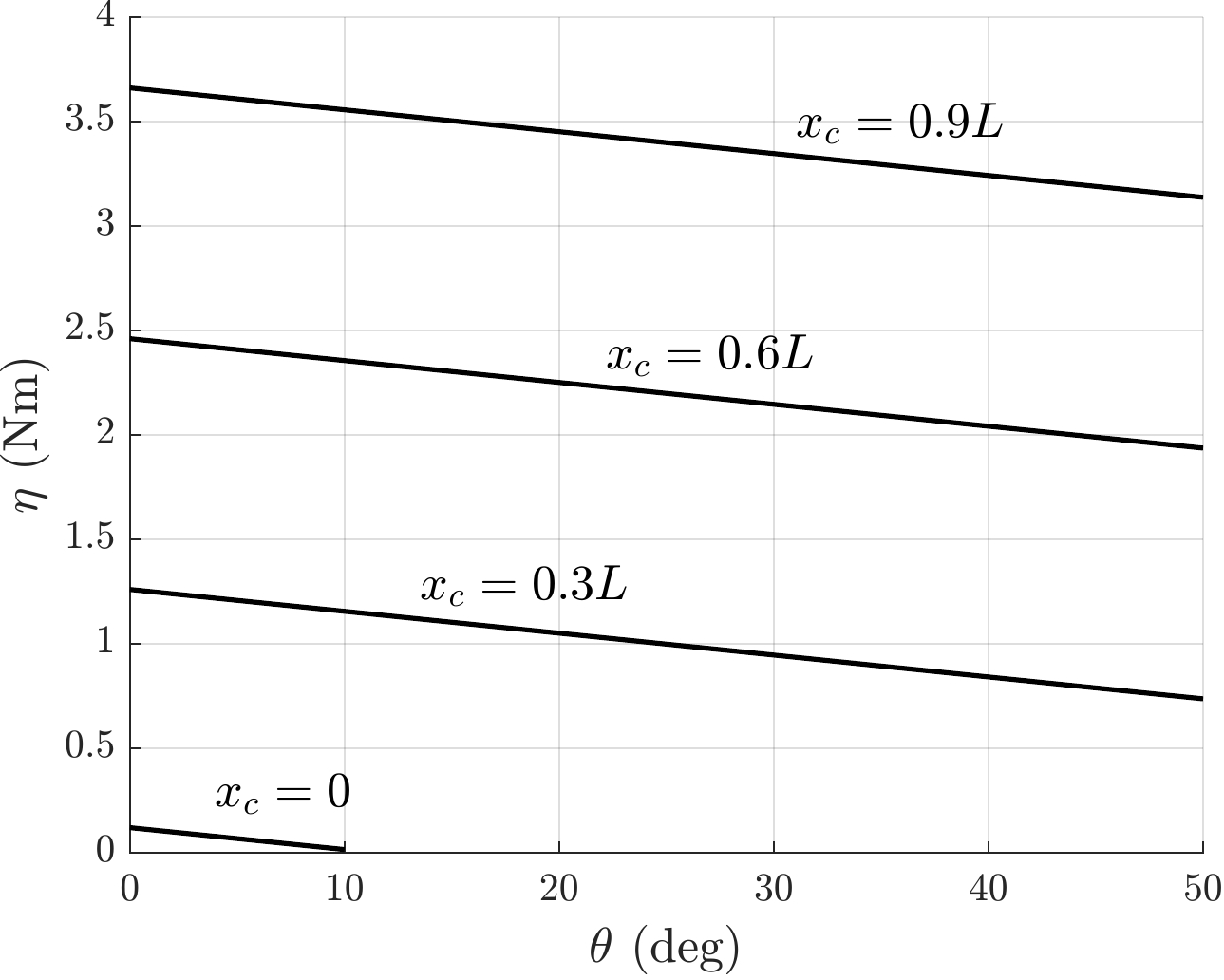}
\caption{Magnitude of the grasp metric with respect to the angle of the door handle for different positions of a pair of antipodal contact points (the computation time required to find $\eta$ at each $x_c$ is about 0.5 seconds).}
\label{Fig:DoorHandle_eta}
\end{figure}

The task-dependent grasp metric $\eta$ in this example is the magnitude of the maximum moment that the grasp can provide about the screw axis $\mathcal{S}$ passing through the handle hinge (Fig.~\ref{Fig:DoorHandle}). In this task, $\bm{l}$ is a unit vector along the axis $\mathcal{S}$ and toward the opposite direction of $z$-axis, and $h=\infty$. Figure~\ref{Fig:DoorHandle_eta} represents the grasp metric $\eta$ with respect to $\theta$ for different positions of the antipodal contact points on the door handle along the $x$-axis ($x_c$). Obviously, by increasing $x_c$, the grasp is able to generate larger moment along the axis $\mathcal{S}$ to turn the door handle. Moreover, while turning the handle, since the require torque increases, the quality of the grasp decreases, such that, for $x_c = 0$, the grasp can turn the handle by only $10^\circ$. It is worthwhile to mention that for $x_c > 0$, the contact $\{c_2\}$ does not contribute to provide the moment about the axis $\mathcal{S}$, however, for $x_c = 0$, both contacts have equal contributions, similar to the task of turning a door knob by a pair of antipodal contact points.


\subsection{Pivoting and Sliding an Object on a Support Surface}
\begin{figure}[!htbp]
\centering
\includegraphics[scale=0.5]{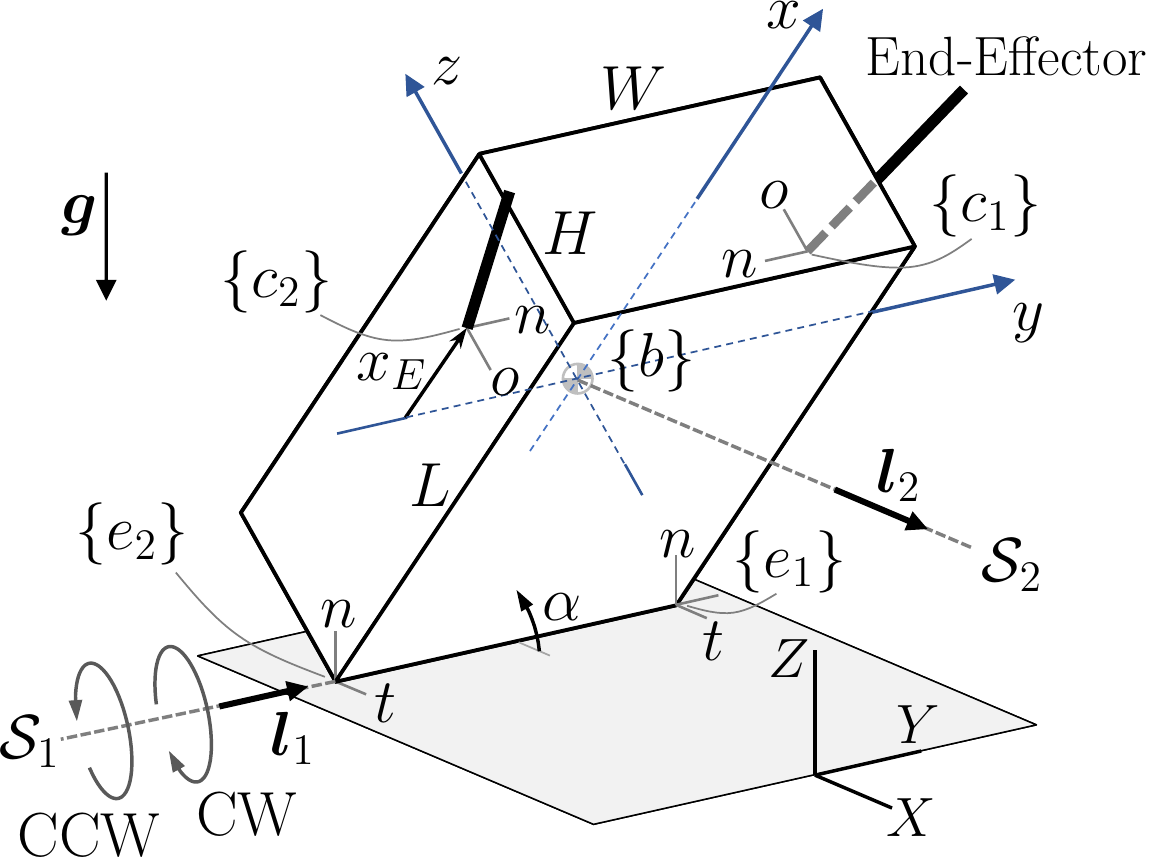}
\caption{Pivoting/sliding a cuboid-shape object on a support surface using two manipulators.}
\label{Fig:PivotingCuboid}
\end{figure}

In this example, we compute the grasp metric for dual-handed quasi-static manipulation of a cuboid-shape object of uniform density by exploiting environment as shown in Fig.~\ref{Fig:PivotingCuboid}. We define two different tasks; (1) pivoting the object about the screw axis $\mathcal{S}_1$ which passes through the edge in contact with the environment, (2) sliding the object along the the screw axis $\mathcal{S}_2$ which passes through the object center of mass. The object-manipulator contacts ($\{c_1\}$, $\{c_2\}$) are considered to be SFCE, the edge contact is modeled by two point contacts with friction (PCWF) at the vertices of the cuboid ($\{e_1\}$, $\{e_2\}$), the body frame $\{b\}$ is attached to the object centroid, and the object weight is the only external wrench applied to the object. Similar to object-manipulator contacts, the environment contact wrenches $\bm{f}_{e_j} =[{f}_{e_{t,j}}, {f}_{e_{o,j}}, {f}_{e_{n,j}}, 0, 0, 0]^\mathrm{T} \in \mathbb{R}^6$ ($j=1,2$) should also satisfy the friction cone constraint $\bm{f}_{e_j} \in {\mathcal{B}_e}_j$ where
\begin{equation}
  {\mathcal{B}_e}_j = \left\{ {\bm{x} \in \mathbb{R}^6} \;\middle|\; \frac{1}{{\mu_e}_j} \sqrt { {\frac{x_1^2}{{e}_{e_{t,j}}^2} } + { {\frac{x_2^2}{{e}_{e_{o,j}}^2}} }} \le x_3, x_{4,5,6}=0 \right\},
  \label{eq:FrictionConeExampleB}
\end{equation}
and ${\mu_e}_j$, ${e}_{e_{t,j}}$, ${e}_{e_{o,j}} \in \mathbb{R}_{++}$ are the parameters of anisotropic friction at the $j$-th environment contact \cite{Patankar2020}.
The simulation parameters are given in Table~\ref{Table:cuboid_parameters} and it is assumed that the maximum normal forces at the object-manipulator contacts are ${f}_{c_{n,1}} = 25 \mathrm{N}$ and ${f}_{c_{n,2}} = 30 \mathrm{N}$.

\begin{table}[!htbp]
    \caption{Simulation parameters for pivoting/sliding the object.}
    \centering
    \setlength{\tabcolsep}{1mm}
    \renewcommand{\arraystretch}{1.2}
    \begin{tabular}{c|c}
        \hline
        \textbf{Parameter}&\textbf{Value}  \\
        \hline 
        \hline
        Weight & $9.81$ (N) \\
        \hline 
        Dimensions & $L = 0.3$ (m), $W = 0.2$ (m), $H = 0.1$ (m)\\
        \hline
        Constants & \begin{tabular}{@{}c@{}} ${\mu_e}_{1,2}=0.25$, ${\mu_c}_{1,2}=0.15$, ${e}_{c_{n,1,2}}=0.06$ (m) \\ ${e}_{c_{t,1,2}}={e}_{c_{o,1,2}}={e}_{e_{t,1,2}}={e}_{e_{o,1,2}}=1$\end{tabular}  \\
        \hline
    \end{tabular}
    \label{Table:cuboid_parameters}
\end{table}

\begin{figure}[!htbp]
\centering
\includegraphics[scale=0.55]{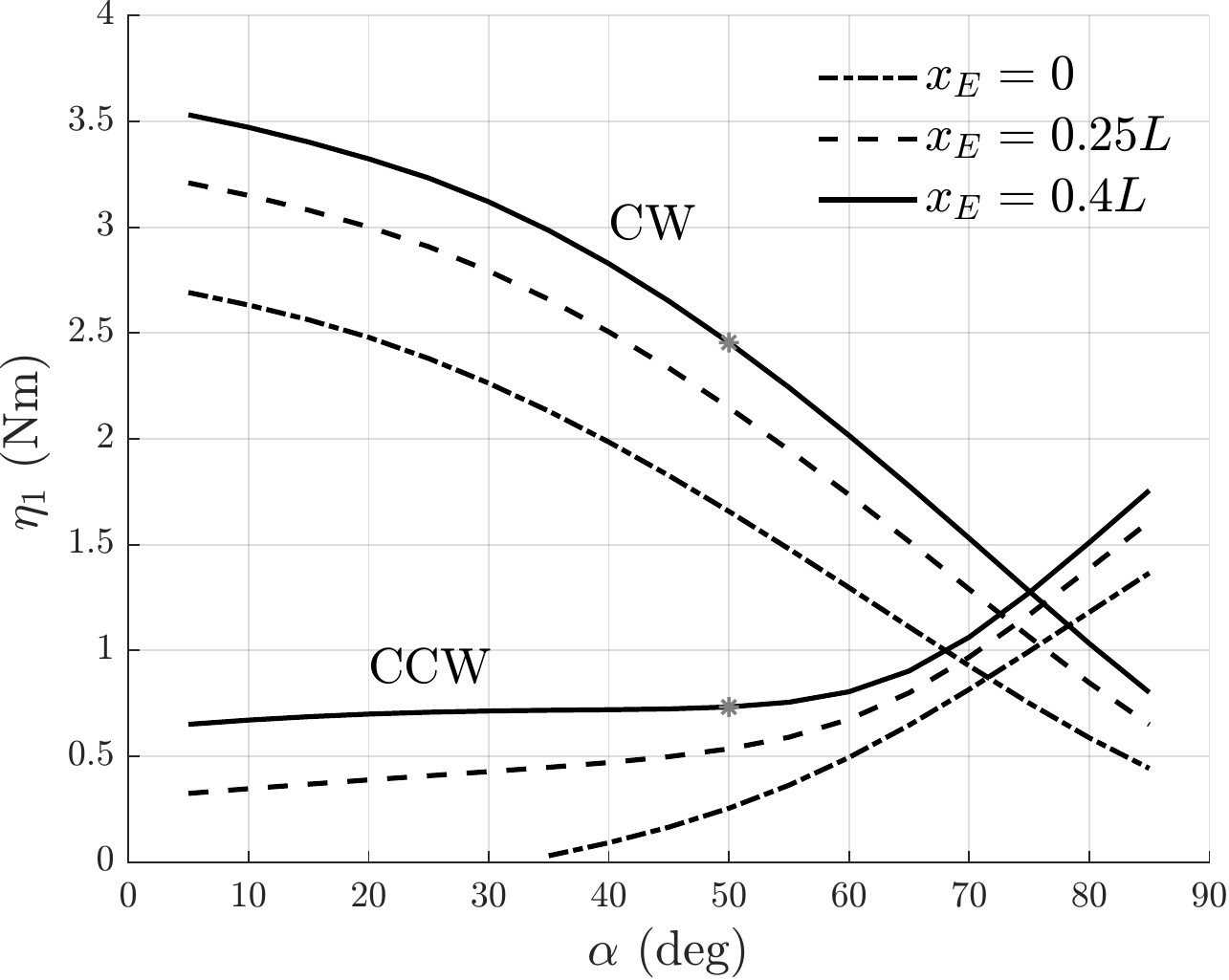}
\caption{Magnitude of the grasp metric about the screw axis $\mathcal{S}_1$ in both clockwise (CW) and counterclockwise (CCW) directions for different rotation angles of the object and three different positions of the end-effectors (the computation time required to find $\eta_1$ at each $\alpha$ is about 0.5 seconds).}
\label{Fig:PivotingCuboid_eta}
\end{figure}

\textbf{Pivoting Task}: The grasp metric $\eta_1$ in this task is the magnitude of the maximum moment that the grasp (including the manipulators, environment, and external wrenches) can provide about the screw axis $\mathcal{S}_1$ in both clockwise (CW) and counterclockwise (CCW) directions (Fig.~\ref{Fig:PivotingCuboid}). In this task, $\bm{l}_1$ is a unit vector along the axis $\mathcal{S}_1$ and toward $y$-axis, and $h_1=\infty$. Figure~\ref{Fig:PivotingCuboid_eta} demonstrates the grasp metric $\eta_1$ for different rotation angles of the object from the supporting surface ($\alpha$) and three different positions of the end-effectors along the $x$-axis from the centroid ($x_E$). The results confirm that due to the effect of the object weight, the metric $\eta_1$ changes during pivoting the object. Moreover, by increasing the distance between the position of the object-manipulator contacts ($\{c_1\}$, $\{c_2\}$) and the axis $\mathcal{S}_1$, the grasp is able to resist a larger disturbance torque along the axis $\mathcal{S}$ in both CW (moving down) and CCW (moving up) directions. Furthermore, the results reveal that by increasing the angle $\alpha$, the maximum disturbance torque the grasp can resist along the axis $\mathcal{S}_1$ in the CCW direction decreases (because the maximum moment the grasp can provide in CW direction decreases) and it increases in the CW direction, and this is because of the effect of the moment of the object weight about the axis $\mathcal{S}_1$.

\begin{figure}[!htbp]
\centering
\includegraphics[scale=0.55]{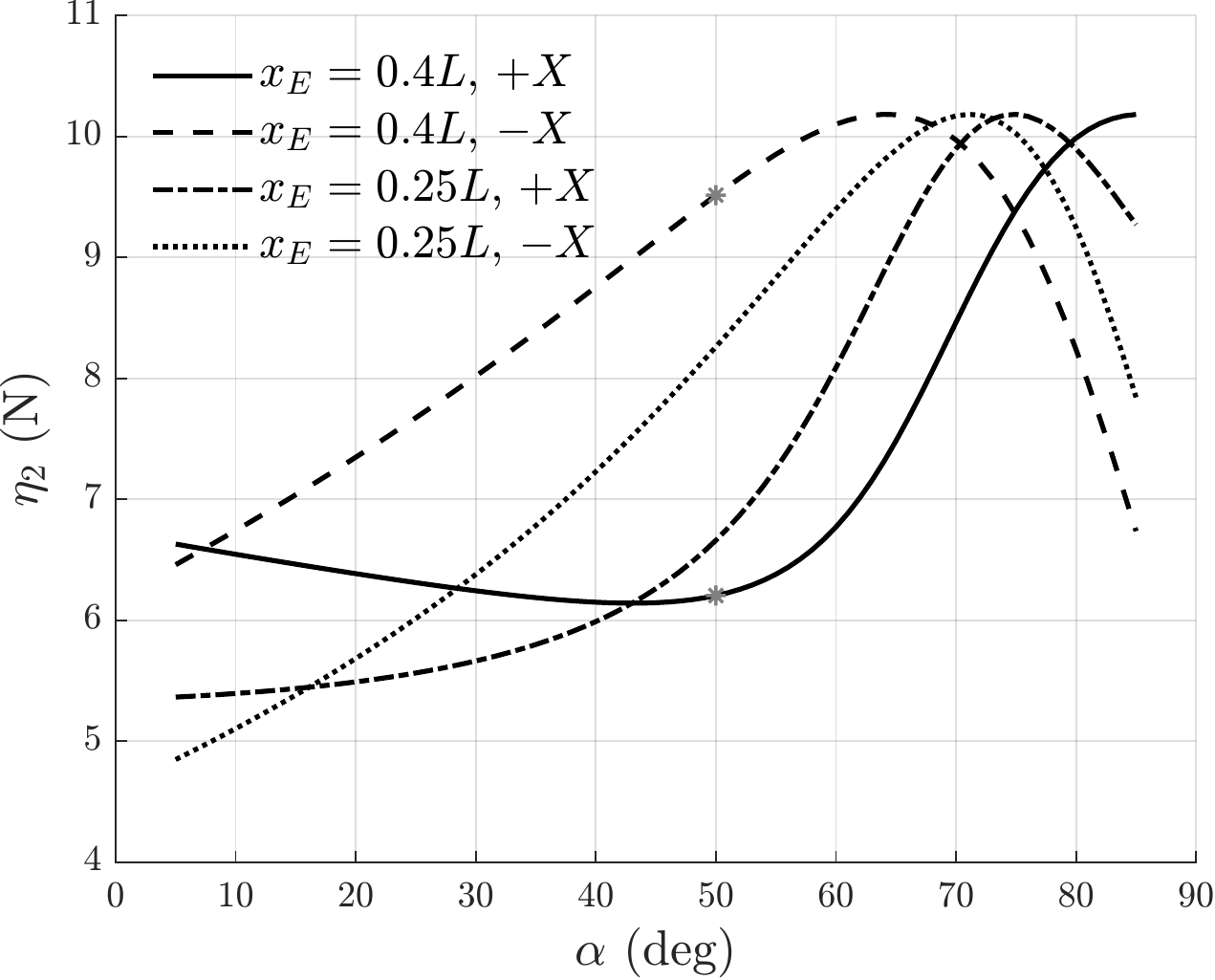}
\caption{Magnitude of the grasp metric along the screw axis $\mathcal{S}_2$ in both $+X$ and $-X$ directions for different rotation angles of the object and two different positions of the end-effectors.}
\label{Fig:SlidingCuboid_eta}
\end{figure}

\textbf{Sliding Task}: The grasp metric $\eta_2$ in this task is the magnitude of the maximum force that the grasp can generate along the screw axis $\mathcal{S}_2$ in both $+X$ and $-X$ directions (Fig.~\ref{Fig:PivotingCuboid}). In this task, $\bm{l}_2$ is a unit vector along the axis $\mathcal{S}_2$ and toward $X$-axis, $h_2=0$, and $\bm{q}_2 = \bm{0}$. In Fig.~\ref{Fig:SlidingCuboid_eta}, the grasp metric $\eta_2$ with respect to the angle $\alpha$ for two different positions of the end-effectors is shown. For a specific grasp, say $\alpha = 50^\circ$ and $x_E = 0.4L$, the maximum disturbance force (exerted at the object center of mass) that the grasp can resist in $+X$ direction (i.e., against pushing) is larger than the maximum disturbance force that the grasp can resist in $-X$ direction (i.e., against pulling). This is because the disturbance force in $-X$ direction results in lower normal forces at the environment contacts ($\{e_1\}$,$\{e_2\}$) and consequently, lower tangential frictional forces (to satisfy the friction cone constraints \eqref{eq:FrictionConeExampleB}), compare with the case the disturbance force in $+X$ direction.

\begin{figure}[!htbp]
    \centering
    \subfloat[]{\includegraphics[scale=0.55]{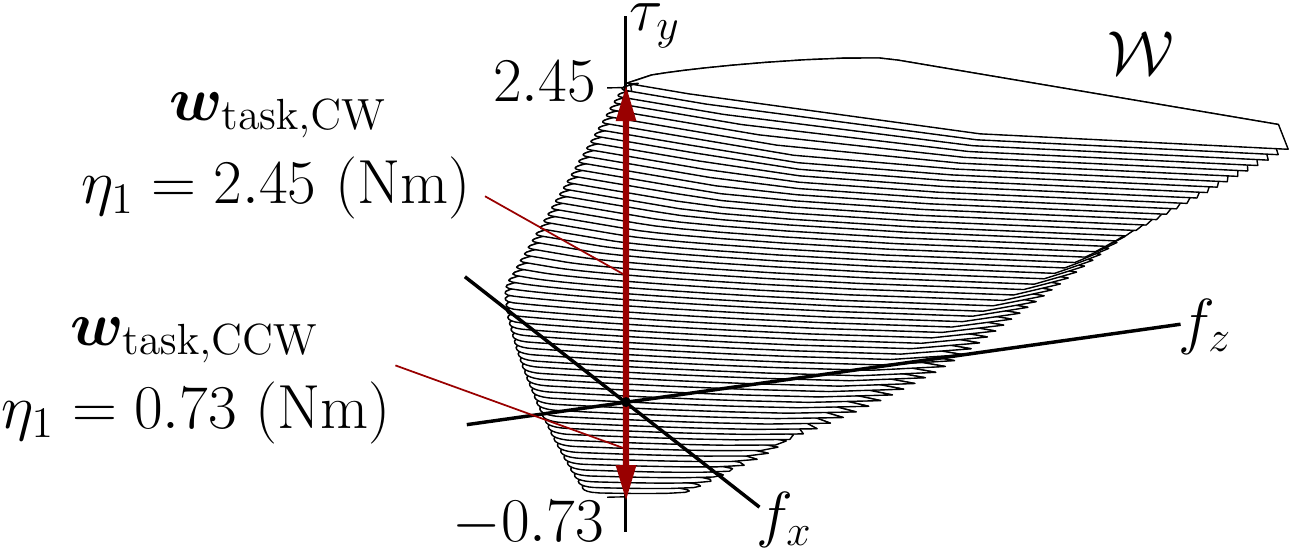}}\\
    \subfloat[]{\includegraphics[scale=0.4]{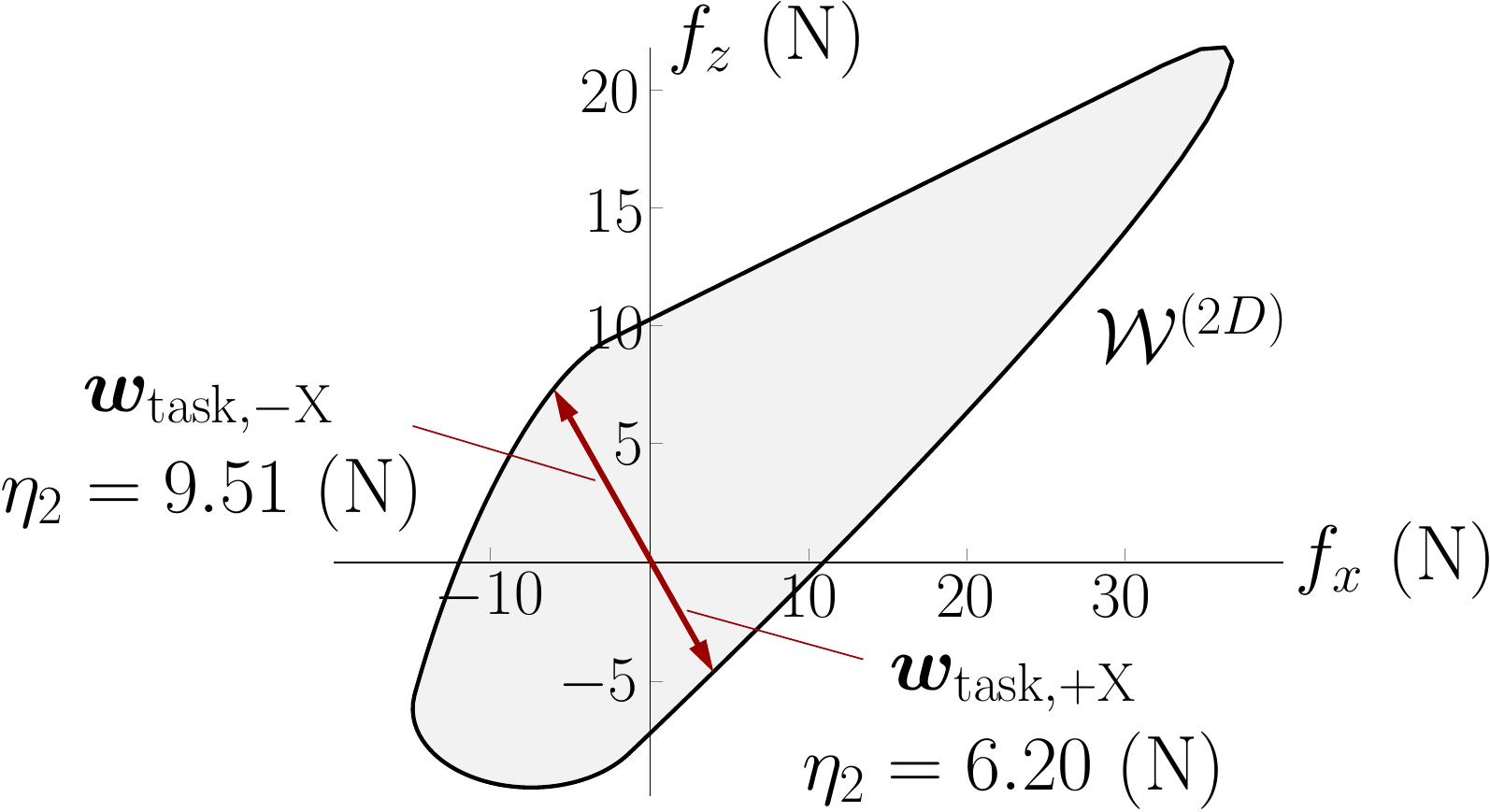}}
    \caption{Grasp wrench space for the grasped object when $\alpha = 50^\circ$ and $x_E = 0.4L$, (a) 3D space, (b) 2D space where $\tau_y = 0$.}
\label{Fig:GWS_Example}
\end{figure}

To have a better understanding of the metrics, the grasp wrench space $\mathcal{W}$ and the maximum wrenches the grasp can provide in both tasks (along $\mathcal{S}_1$ and $\mathcal{S}_2$) for a particular case of $\alpha = 50^\circ$ and $x_E = 0.4L$ are shown in Fig.~\ref{Fig:GWS_Example}. The magnitudes of these wrenches represent the grasp metrics given in Fig.~\ref{Fig:PivotingCuboid_eta} and Fig.~\ref{Fig:SlidingCuboid_eta}. Note that the GWS, in general, is a six-dimensional space, however, in this example the task wrenches lie only in three dimensions of $f_x$, $f_z$, and $\tau_y$ \cite{MillerAllen1999}.



In almost all the metrics proposed in the literature, it is assumed that the upper limit of the magnitude of the forces \cite{Borst2004,pollard1994parallel,haschke2005task} or the normal forces \cite{ZhengQian2009} applied to the object at the contacts are equal. Moreover, the object weight is ignored or considered as part of the task \cite{li1988task},\cite{song2020robust}. However, this example reveals the power of our formulation in determining a task-dependent metric that can deal with the different contact force upper limits,
and separating the object weight from an specific task.

\section{Conclusion and Future Work}
\label{sec:conc}
We present a second-order cone program (SOCP) to compute a task-dependent grasp metric where the task is defined by a task screw (a unit magnitude screw about which the grasp needs to apply a wrench to generate a desired motion). In contrast to other methods based on unrealistic assumptions of bounding all the contact forces by the same value or bounds some norm (either $1$, $2$ or $\infty$) of the vector of concatenated contact forces, our formulation includes independent lower and upper bounds on the normal contact forces of manipulators and environment, which allows the maximum allowable contact force at each individual contact to be different.  Moreover, the SOCP directly incorporates the friction cone constraints at each contact as second-order cone constraint, which distinguishes our approach from other methods that approximates the friction cone with a polytope or models it as a linear matrix inequality by embedding it in a higher dimensional space. 
We evaluate the grasp metric in three tasks of manipulating with the help of the environment contact, namely, turning a door handle, pivoting an object, and sliding an object on a support surface. 
For each task, for a given screw motion to accomplish the task, we can compute the metric for each grasp (with given contact positions of the manipulators and the configuration of the object). 

As also noted in~\cite{haschke2005task}, the key limitation of our task definition is that our grasp metric, cannot guarantee the robustness to unknown disturbances. We plan to further investigate this in future work.  For future work, we also plan to use our grasp evaluation method to design grasp planning algorithms especially for manipulation tasks where the environment contact can be exploited (e.g., pivoting a cuboid object about it's corners on a support plane).


\section*{APPENDIX}
In this section, the definition of some of the key terminology used in the paper are presented.

\noindent
$\bullet$ A \textbf{Screw Motion} is a rotation $\theta \in \mathbb{R}$ about an axis $\bm{l} \in \mathbb{R}^3$ along with a translation $d \in \mathbb{R}$ parallel to this axis. $M = \theta$ and $h = d/\theta$ are known as magnitude and pitch of the screw motion, respectively (if $\theta = 0$, then $h=\infty$, $M = d$).

\noindent
$\bullet$ A \textbf{Twist} $\bm{\xi} = [\bm{v}^\mathrm{T},\bm{\omega}^\mathrm{T}]^\mathrm{T} \in \mathbb{R}^6$  is the instantaneous velocity of a rigid body in terms of its linear $\bm{v} \in \mathbb{R}^3$ and angular $\bm{\omega} \in \mathbb{R}^3$ components.

\noindent
$\bullet$ A \textbf{Screw Associated with a Twist} $\bm{\xi}$ represents the screw coordinates of the twist as pitch $h = \bm{\omega}^\mathrm{T} \bm{v} / {\lVert \bm{\omega} \rVert}^2 \in \mathbb{R}$, axis $\bm{l} = \bm{\omega} \times \bm{v} / {\lVert \bm{\omega} \rVert}^2 + \lambda \bm{\omega} \in \mathbb{R}^3$ ($\forall\lambda \in \mathbb{R}$), and magnitude $ M  = {\lVert \bm{\omega} \rVert} \in \mathbb{R}$ (if $\bm{\omega} = \bm{0}$, then $h=\infty$, $\bm{l} =\lambda \bm{v}$, and $M = {\lVert \bm{v} \rVert}$).

\noindent
$\bullet$ A \textbf{Wrench} $\bm{w} = [\bm{f}^\mathrm{T},\bm{\tau}^\mathrm{T}]^\mathrm{T} \in \mathbb{R}^6$ is the generalized force acting at a point on a rigid body, which consists of the force $\bm{f} \in \mathbb{R}^3$ and moment $\bm{\tau} \in \mathbb{R}^3$ components.

\noindent
$\bullet$ A \textbf{Screw Associated with a Wrench} $\bm{w}$ represents the screw coordinates of the wrench as pitch $h = \bm{f}^\mathrm{T} \bm{\tau} / {\lVert \bm{f} \rVert}^2 \in \mathbb{R}$, axis $\bm{l} = \bm{f} \times \bm{\tau} / {\lVert \bm{f} \rVert}^2 + \lambda \bm{f} \in \mathbb{R}^3$ ($\forall\lambda \in \mathbb{R}$), and magnitude $ M  = {\lVert \bm{f} \rVert} \in \mathbb{R}$ (if $\bm{f} = \bm{0}$, then $h=\infty$, $\bm{l} =\lambda \bm{\tau}$, and $M = {\lVert \bm{\tau} \rVert}$).


\noindent
$\bullet$ A \textbf{Constant Screw Motion} is a motion where only the magnitude of the screw associated with the twist changes, not the pitch and the axis.

\noindent
$\bullet$ A \textbf{Constant Screw Wrench} is a wrench where only the magnitude of the screw associated with the wrench changes, not the pitch and the axis.

\noindent
$\bullet$ A \textbf{Second Order Cone Program (SOCP)} is a special class of convex optimization problems of the form
\begin{equation}
\begin{aligned}
& {\underset {\boldsymbol{x}}{\text{maximize} }} & & \boldsymbol{f}^{T} \boldsymbol{x} \\
& \text{subject to}  & &  \left\|\bm{\mathrm{A}}_{i} \boldsymbol{x} +\boldsymbol{b}_{i}\right\|_{2} \leq \boldsymbol{c}_{i}^{T} \boldsymbol{x} + d_{i}, \,\, i=1,..., m, \\
&&& \bm{\mathrm{F}} \boldsymbol{x} = \boldsymbol{g},
\end{aligned}
\label{equation:SOCP}
\end{equation}
where $ \boldsymbol{x}\in \mathbb {R} ^{n}$ is the optimization variable, $\left\|\cdot\right\|_{2}$ is the Euclidean norm, and $ \boldsymbol{f}\in {\mathbb  {R}}^{n}$, $\bm{\mathrm{A}}_{i}\in {\mathbb  {R}}^{{{n_{i}}\times n}}$, $\boldsymbol{b}_{i}\in {\mathbb  {R}}^{{n_{i}}}$, $\boldsymbol{c}_{i}\in {\mathbb  {R}}^{n}$,  $d_{i}\in {\mathbb  {R}}$, $\bm{\mathrm{F}}\in {\mathbb  {R}}^{{p\times n}}$, $\boldsymbol{g}\in \mathbb {R} ^{p}$ are the problem parameters \cite{boyd2004convex}.


\bibliographystyle{IEEEtran}
\bibliography{References}

\end{document}